\documentclass{article}

\usepackage{subcaption}

\usepackage{arxiv}

\usepackage[utf8]{inputenc} 
\usepackage[T1]{fontenc}    

\usepackage{url}            
\usepackage{booktabs}       
\usepackage{amsfonts}       
\usepackage{nicefrac}       
\usepackage{microtype}      
\usepackage{graphicx}
\usepackage[square,numbers]{natbib}
\usepackage{wrapfig}

\usepackage{doi}

\usepackage{amsmath}
\usepackage{tabularx}
\usepackage{multicol}
\usepackage{multirow}
\usepackage{caption}
\usepackage{hhline}
\usepackage{bbding}
\usepackage{makecell}
\usepackage{array}
\usepackage{bm}
\usepackage{relsize}

\newcolumntype{?}{!{\vrule width 1pt}}

\newcommand\Tstrut{\rule{0pt}{2.6ex}}       
\newcommand\Bstrut{\rule[-1.5ex]{0pt}{0pt}} 
\newcommand{\TBstrut}{\Tstrut\Bstrut}
\newcommand\freefootnote[1]{%
  \let\thefootnote\relax%
  \footnotetext{#1}%
  \let\thefootnote\svthefootnote%
}

\author{Ali Abbasi\\
Vanderbilt University\\
{\tt\small ali.abbasi@vanderbilt.edu}
\And
Parsa Nooralinejad\\
University of California, Davis\\
{\tt\small pnoorali@ucdavis.edu}
\And
Hamed Pirsiavash\\
University of California, Davis\\
{\tt\small hpirsiav@ucdavis.edu}
\And
Soheil Kolouri\\
Vanderbilt University\\
{\tt\small soheil.kolouri@vanderbilt.edu}
}

\title{BrainWash: A Poisoning Attack to Forget in Continual Learning}

\date{}

\hypersetup{
pdftitle={BRAINWASH: A POISONING ATTACK TO FORGET IN CONTINUAL
LEARNING},
pdfauthor={Ali Abbasi, Parsa Nooralinejad, Hamed Pirsiavash, Soheil Kolouri},
pdfkeywords={Continual Learning, Machine Learning},
}

\DeclareMathOperator*{\argmax}{arg\,max}
\DeclareMathOperator*{\argmin}{arg\,min}

\begin{document}
\maketitle

\begin{abstract}
	Continual learning has gained substantial attention within the deep learning community, offering promising solutions to the challenging problem of sequential learning. Yet, a largely unexplored facet of this paradigm is its susceptibility to adversarial attacks, especially with the aim of inducing forgetting. In this paper, we introduce "BrainWash," a novel data poisoning method tailored to impose forgetting on a continual learner. By adding the BrainWash noise to a variety of baselines, we demonstrate how a trained continual learner can be induced to forget its previously learned tasks catastrophically, even when using these continual learning baselines. An important feature of our approach is that the attacker requires no access to previous tasks' data and is armed merely with the model's current parameters and the data belonging to the most recent task. Our extensive experiments highlight the efficacy of BrainWash, showcasing degradation in performance across various regularization-based continual learning methods.
\end{abstract}

\freefootnote{Preprint.}


\section{Introduction}
\label{sec:intro}

In real-world scenarios, data distributions are inherently non-stationary, constantly evolving and shifting in unpredictable ways. Such variability poses a significant challenge to machine learning and computer vision, where model generalizability assumes stationary training and testing/deployment distributions. Continual Learning (CL) \cite{de2021continual,kudithipudi2022biological} has emerged as a prolific research domain focusing on efficient learning from an ongoing stream of data or tasks. CL primarily seeks to: 1) enhance backward knowledge transfer, which aims to maintain or improve performance on previously learned tasks, thereby mitigating catastrophic forgetting, and 2) bolster forward knowledge transfer, where learning a current task can boost performance on or reduce the learning time for future tasks.  CL has significantly progressed in computer vision tasks, including incremental image recognition \cite{van2020brain,kim2023ancl}. With the increase in the adoption of CL algorithms, examining their vulnerabilities is imperative to inform the development of more robust CL methodologies.

Most research in Continual Learning (CL) has focused on developing algorithms to maximize backward and forward knowledge transfer. Existing methods can generally be categorized into three groups: 1) replay-based methods, 2) regularization-based methods, and 3) parameter isolation methods. Nonetheless, there has been limited focus on the robustness of CL approaches against various types of adversarial attacks. Recent studies have begun to address this gap by proposing backdoor attacks \cite{umer2020targeted,kang23poisoning} and certain poisoning attacks \cite{li2022targeted,han2023data} within the CL context. These contributions are critical in profiling the vulnerabilities of CL methods, paving the way for developing more resilient CL algorithms. Additionally, these findings have implications for closely related and emerging fields such as machine unlearning.

\begin{figure}[t!]
    \centering \includegraphics[width=0.7\columnwidth]{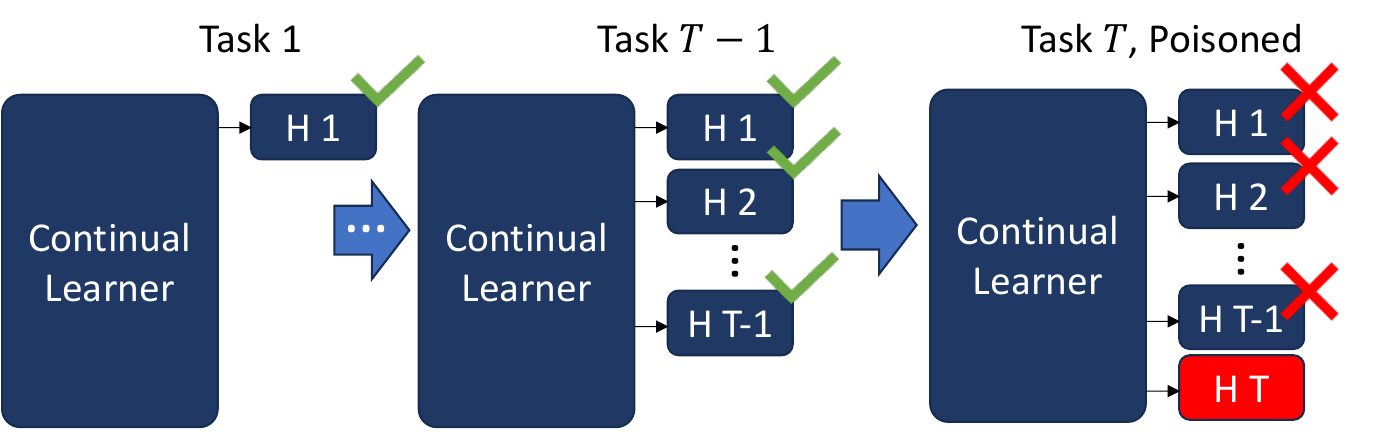}
    \caption{BrainWash is a poisoning attack targeting continual learning systems. It sabotages a task so that, upon learning it, the system's rate of forgetting previously learned tasks is increased.}
    \label{fig:teaser}
\end{figure}

Recent works have confirmed that an adversary can insert misinformation into a task to distort a continual learner's performance. For instance, Umer et al.  \cite{umer2020targeted} show that backdoors can be placed into a task to hijack the performance of a CL method, and the backdoor remains effective even when new tasks are learned.  Here, we pose a fundamental question: Is it possible to `brainwash' a continual learner by poisoning its current task in such a way that performance on all previous tasks is significantly degraded? More succinctly, can a task be designed to induce maximum forgetting of prior knowledge in a CL context? We affirmatively answer this question and demonstrate its validity across a wide range of regularization-based CL methods, assuming minimal and realistic conditions. This concept is depicted in Figure \ref{fig:teaser}.

Recent advancements in foundational models have given rise to massive models boasting billions of parameters. These models, essential for tasks like training future Large Language Models (LLMs) on extensive datasets, demand substantial data resources. However, the computational power required for training is constrained, preventing multiple passes over the training data. Complicating matters further is that the data isn't sampled independently and identically distributed (i.i.d.), as continual improvement of already trained models necessitates incorporating recently available web data. In this context, foundational models must employ continual learning methods to avoid forgetting previously acquired knowledge. This vulnerability to forgetting poses a risk, as adversaries can exploit it by introducing subtly manipulated new training data to induce the model to forget key information.


In this paper, we examine a realistic threat model targeting regularization-based continual learning methods. Under this model, the attacker has access only to the victim's model and knowledge of the subsequent task that the victim will tackle. Crucially, the attacker remains unaware of the specific CL algorithm employed by the victim to learn tasks and lacks access to data from prior tasks. We propose a novel method denoted as ``BrainWash'' that allows for poisoning the current task data to maximize forgetting on prior tasks. 

\begin{figure}[t!]
    \hspace*{-10pt} \includegraphics[width=\columnwidth]{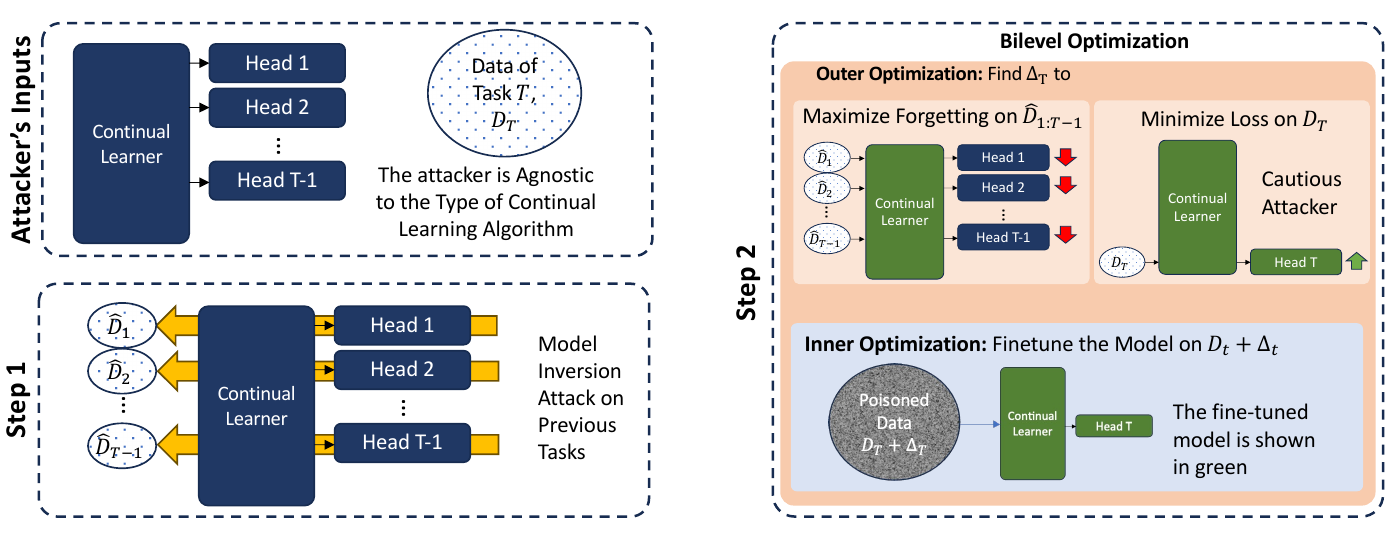}
    \caption{In our proposed threat model, the attacker gains access to the Continual Learning (CL) model and the data for the forthcoming task but remains unaware of the data from preceding tasks and the specific CL method employed by the victim (top panel). The attack methodology unfolds in two steps. Firstly, the attacker executes a model inversion attack on the CL model to reconstruct an approximation of the victim's data from earlier tasks (middle panel). Secondly, the attacker employs bi-level optimization to contaminate the data for the current task. This is done in such a way that performance on the reconstructed data from previous tasks is significantly degraded.}
    \label{fig:overview}
\end{figure}

In short, BrainWash consists of two main steps. First, we perform a model inversion attack \cite{fredrikson2015model,veale2018algorithms} on the continual learner to approximate the data from the previous tasks. Second, to poison the current task, we construct a bi-level optimization problem such that: 1) the performance on inverted data of previous tasks is minimized, and 2) the performance on the clean data of the current task is maximized. Figure \ref{fig:overview} demonstrates the threat model as well as the two steps of BrainWash.

\noindent \textbf{Contributions.} Our main contributions in this paper are: 

\begin{enumerate}
    \item Devising a novel poisoning attack algorithm for regularization-based continual learning methods, denoted as BrainWash. 
    \item Demonstrating the effectiveness of BrainWash on benchmark CL datasets and across diverse regularization-based CL algorithms. 
    \item Providing extensive ablation studies to deepen our understanding of BrainWash.
\end{enumerate}

\section{Related Work}
\textbf{Continual Learning} is a subfield of ML focused on learning from nonstationary streams of data or tasks \cite{de2021continual,kudithipudi2022biological}. Its objectives include improving backward knowledge transfer to maintain or enhance performance on previously learned tasks helping to prevent catastrophic forgetting. It also aims to strengthen forward knowledge transfer, where mastering a current task can improve performance or decrease learning time for future tasks. Catastrophic forgetting prevention is a central goal in this field. To tackle catastrophic forgetting, strategies in continual learning are typically grouped into three main categories: 1) memory-based methods, 2) regularization-based methods, and 3) architectural methods.  Memory-based methods involve techniques such as memory rehearsal or replay, generative replay, and gradient projection \citep{shin2017continual,farquhar2018towards,van2018generative,van2020brain,rolnick2019experience,rostami2020generative,farajtabar2020orthogonal,saha2020gradient,wang2021training,lin2022trgp,abbasi2022sparsity}. These methods often rely on storing and revisiting previous learning experiences or artificially generating them to reinforce learning. Regularization-based methods apply penalties on changing parameters that are vital for tasks already learned \citep{kirkpatrick2017overcoming,zenke2017continual,aljundi2017memory,kolouri2020sliced,li2021lifelong,von2019continual}. These approaches help in preserving the knowledge acquired from previous tasks while allowing new learning. Architectural methods focus on modifying the learning model itself. Strategies include expanding the model structure \citep{Rusu2016,schwarz2018progress}, isolating parameters specific to certain tasks \citep{mallya2018packnet,mallya2018piggyback}, and using masking techniques \cite{wortsman2020supermasks,ben-iwhiwhu2023lifelong,nath2023L2D2} to manage the learning process for different tasks. In this paper, we focus on regularization-based methods, mainly due to their effective balance between plasticity and stability, allowing for the integration of new knowledge while preserving essential information from past learning experiences. We propose a data poisoning attack that maximizes forgetting for regularization-based continual learners.

\textbf{Data Poisoning} is a training phase attack on a machine learning model in which the attacker deliberately alters the victim's training data maliciously \cite{biggio2012poisoning,alfeld2016data,jagielski2018manipulating,shafahi2018poison,feng2019deep,zhu2019transferable,geiping2021witches}. After the victim trains their model using this compromised data, the model would serve the attacker's detrimental objectives, such as significantly reducing the model's test accuracy on all or specific classes (i.e., targeted vs. non-targeted attacks). 

Data poisoning is formally defined as a bi-level optimization problem \cite{bard1982explicit,biggio2012poisoning}. In the outer level optimization, the attacker optimizes the poisoning, which can be additive noise \cite{geiping2021witches}, patch-based noise \cite{chen2021deeppoison}, or a conditional generative model for noise \cite{feng2019deepconfuse}, to enforce their malicious intention on the `resulting network' parameters. This `resulting network' itself is the solution to the inner optimization problem that minimizes the training objective as would be done by the victim. When the ML model is a deep neural network, this bi-level optimization problem is generally intractable, as it requires backpropagation through the entire SGD training procedure \cite{munoz2017towards}. Hence, the existing literature often approximates this bi-level optimization using various strategies, including first-order approximation methods \cite{huang2020metapoison} and more sophisticated methods based on alternating optimization \cite{feng2019deepconfuse}. Similar to \cite{finn2017mamal,huang2020metapoison}, our poisoning attack also uses a first-order approximation method for solving the induced bi-level optimization. In contrast with \cite{huang2020metapoison}, however, our bi-level optimization objective is maximizing forgetting in a continual learner. 

\noindent \textbf{Model Inversion} \cite{fredrikson2015model,veale2018algorithms,nguyen2023re} encompasses attack strategies designed to either reconstruct training data or deduce sensitive attributes from a trained model. These strategies are broadly divided into `optimization-based' and `training-based' methods. Our study primarily explores optimization-based methods, which are widely adopted in the literature \cite{yin2020dreaming,nguyen2023re}. These methods primarily adjust inputs in the data space to maximally stimulate specific output neurons, such as target classes. However, a key challenge arises from the many-to-one mapping characteristic of deep neural networks, where a variety of inputs can lead to the same output. To address this, the literature introduces various forms of priors or regularization terms, making this optimization process more tractable. Such regularization terms range from simpler approaches like Total Variation and image norm \cite{mahendran2015understanding,mordvintsev2015deepdream} to more advanced techniques involving feature statistics \cite{yin2020dreaming} and the use of generative models \cite{wang2021variational}. In this paper, we adopt a model inversion approach similar to Yin et al. \cite{yin2020dreaming} to approximate the data that the continual learner has been trained on from previous tasks.

\section{Threat model}
We consider a victim using a regularization-based Continual Learning (CL) method to learn a series of tasks. For example, imagine a home robot that continuously learns from its environment, such as adapting to a new home \cite{chang2023goat}. The attacker's objective is to poison the training data of the latest task (like learning about a new room), causing the CL model to forget previously learned tasks upon acquiring new information. Furthermore, the attacker poisons the data in our setup by engineering norm-constrained additive noise. We examine two scenarios for such an attack: 1) the `reckless threat model' and 2) the `cautious threat model'. In the `reckless threat model', the victim is assumed to deploy the model without evaluating/monitoring its performance on the latest task. Thus, the attacker focuses on maximizing forgetting of prior tasks without concern for the performance of the current task. In contrast, the `cautious threat model' presupposes that the victim monitors the CL model's performance on the latest, potentially poisoned task. A significant drop in performance on this task could raise suspicions. Consequently, in the `cautious threat model', the attacker must carefully balance the attack to not only induce forgetting of previous tasks but also retain acceptable accuracy on the current task, presenting a significantly more challenging setting. In both settings, we assume that the attacker does not have access to the continual learner's training data from previous tasks. 


\section{Method}

In this work, we aim to design a poisoning attack for regularization-based multi-head CL approaches that brainwashes the model, causing it to forget its previous tasks. We assume the attacker has full access to the model and data from the latest task the continual learner will encounter. However, the attacker does not have access to continual learner's data from the previous tasks, as regularization-based CL methods discard the data of earlier tasks. 


We propose to utilize model inversion attacks \cite{fredrikson2015model,yin2020dreaming} to 
obtain an approximation for the continual learner's data from prior tasks. Using the victim's model, the inverted data from previous tasks, and the data for the current task, the attacker formalizes the poisoning problem through a bi-level optimization and then solves it via a first-order approximation method. In what follows, we briefly review our notations and then describe 1) the model inversion attack, 2) poisoning as a bi-level optimization problem, and 3) our proposed first-order approximation solver.


\subsection{Notations}

We denote the training data for task $t\in\{1,\cdots,T\}$ as $D_t=\{(x^i_t,y^i_t)\}_{i=1}^{N_t}\subset \mathcal{X}\times\mathcal{Y}_t$, where $x^i_t\in \mathcal{X}$ denotes the $i$'th sample from the $t$'th task (e.g., an input image) and $y_t^i\in \mathcal{Y}_t=\{1,\cdots,K_t\}$ denotes its corresponding label with $K_t$ and $N_t$ denoting the number of classes and examples for task $t$ respectively. Let $f(\cdot;\theta)$ denote the CL's backbone that extracts deep representations from the input data, where $\theta$ indicates the backbone's parameters, and let $h_t(\cdot;\psi_t)$ denote the classification head for task $t$, with $\psi_t$ representing its parameters.   

Throughout the paper, we consider the supervised classification problem, and denote the classification loss (e.g., cross-entropy) as $\mathcal{L}(\cdot)$. Moreover, we use $\ell_p(\cdot)$ to denote the $p$'th norm of a vector, and in particular, use $\ell_\infty$ norm in our experiments. Lastly, we indicate learned parameters calculated on clean data with a superscript asterisk and those calculated on the poisoned data with a tilde. For instance, \(\psi^*_t\) represents the optimal parameters  for the \(t^{th}\) head, while \(\theta^*_{1:T-1}\) denotes the optimal parameters of the backbone after learning tasks \(1\) to \(T-1\), all calculated on the clean data. And $\tilde{\theta}$ and $\tilde{\psi}_T$ denote the backbone parameters and the parameters of the $T$'th head after poisoning.


    
    

\subsection{Model Inversion}\label{sec:inversion}
We propose executing a model inversion on the victim's CL model to approximate data from previous tasks. The outcome of this attack is proxy datasets for the previous tasks. The outcome of this attack is an approximate dataset $\hat{D}_t=\{(\hat{x}_t^i,\hat{y}_t^i)\}_{i=1}^{M}$ per task, where $\hat{x}_t^i$ is an inverted sample corresponding to label $\hat{y}_t^i$. Below we describe the process to construct $\hat{D}_t$ for the previous tasks, i.e., for $t\in\{1,\cdots,T-1\}$. 

For any past task $t$,  the attacker can infer the number of classes, $K_t$, by examining the logits in the $t$'th head, $h_t(\cdot;\psi^*_t)$. In line with the work of  Yin et al. \cite{yin2020dreaming}, we formulate the model inversion on task $t$ for a set of randomly sampled target one-hot labels $\{\hat{y}_t^i\in \mathcal{Y}_t\}_{i=1}^M$ as:
\begin{align}
\label{eq:inversion}
    \{\hat{x}_t^i\}_{i=1}^M =  \argmin_{\{x^i\in \mathcal{X}\}_{i=1}^M}  \sum_{i=1}^M \mathcal{L}(x^i, \hat{y}_t^i,\theta^*_{1:T-1},\psi^*_t) + 
    \sum_{i=1}^M \mathcal{R}_{\text{prior}}(x^i)
    +\alpha_f \mathcal{R}_{\text{feat}}(\{x^i\}_{i=1}^M, \theta_{1:T-1}^*),\nonumber    
\end{align}
where \(\mathcal{R}_{\text{prior}}\) is an image regularization term that acts as a weak prior for natural images \cite{mordvintsev2015deepdream}, and \(\mathcal{R}_{feat}\) is a feature-statistics regularization as introduced in \cite{yin2020dreaming}. For \(\mathcal{R}_{\text{prior}}(x)\) we use:
\begin{align}
    \mathcal{R}_{\text{prior}}(x) = \alpha_{\text{TV}}\mathcal{R}_{\text{TV}}(x)+ \alpha_{\ell_2}\mathcal{R}_{\ell_2}(x),
\end{align}
where $\mathcal{R}_{TV}(x)$ represents the total variation of image $x$, $\mathcal{R}_{\ell_2}(x)$ is the $\ell_2$ norm of the image, and $\alpha_{\text{TV}},\alpha_{\ell_2},\alpha_f>0$ denote the regularization coefficients. 

Furthermore, the feature-statistics regularization \(\mathcal{R}_{feat}\) leverages the prevalence of batch normalization layers \cite{ioffe2015batch} in modern deep neural networks and the fact that they maintain a running mean and variance of training representations.  Hence,  \(\mathcal{R}_{feat}\) requires the feature-statistics of the inverted samples $\{\hat{x}^i_t\}_{i=1}^M$ to align with those of the batch normalization layers, via:
\begin{align}
\mathcal{R}_{\text{feat}}(\{x^i\}_{i=1}^M,\theta^*_{1:T-1}) = \sum_l \left\| \mu_l(\{x^i\}_{i=1}^M) - m_l) \right\|^2_2 + \sum_l \left\| \sigma_l^2(\{x^i\}_{i=1}^M) - v_l \right\|^2_2.
\end{align}
Here, \(m_l\) and \(v_l\) are the running means and variances stored at the \(l^{\text{{th}}}\) batch normalization layer, and $\mu_l$ and $\sigma^2_l$ are the corresponding mean and variance of $\{x^i\}_{i=1}^M$ across examples at this layer. Note that our model inversion does not rely heavily on this regularizer, so it is still applicable in other architectures with no batch normalization layer.








\subsection{Poisoning Formulation}

As our main contribution, we formalize our poisoning attack for the `reckless' attacker. The attacker aims to construct additive noise to the data from task $T$, such that when the victim trains their model on this task, the performance on tasks $1$ to $T-1$ plummets. Furthermore, we assume that the attacker is oblivious to the victim's specific continual learning approach. Mathematically, we formalize the `reckless' attacker problem as a bi-level optimization problem:
\begin{align}
\{\delta^i_T\}_{i=1}^{N_T} &= \argmax_{\{\delta^i\}_{i=1}^{N_T}} \sum_{t=1}^{T-1}\sum_{j=1}^{M}\mathcal{L}(\hat{x}^j_t, \hat{y}^j_t, \tilde{\theta}(\delta), \psi^*_t) \nonumber \\
s.t. \quad & \tilde{\theta}(\delta), \tilde{\psi}_T(\delta) = \argmin_{\theta, \psi_T} \sum_{i=1}^{N_T}\mathcal{L}(x^i_T + \delta^i, y^i_T, \theta, \psi_T) \nonumber \\
&~~~~~~~~\ell_{\infty} (\delta^i) < \epsilon,~~\forall i
\label{eq:reckless}
\end{align}
Here, $T$ denotes the most recent task, $\delta^{i}_T$ is the optimal additive noise for sample $i$ of task $T$, resulting in the poisoned data $x_T^i+\delta_T^i$,  $\hat{x}^j_t$ represents the $j$'th inverted sample from task $t$, and $\epsilon$ is the threshold for the $\ell_\infty$ norm of the noise, which ensures inconspicuousness. Lastly, $\tilde{\theta}(\delta)$ is the updated parameters of the CL model when trained on the poisoned data from task $T$, which depends on the noise, $\delta$. Importantly, in the inner optimization, the attacker is simply fine-tuning the CL model on the poisoned data of task $T$, starting from $\theta^*_{1:T-1}$ and ending at $\tilde{\theta}(\delta)$. The attacker's fine-tuning in the inner optimization would differ from the victim's learning of task $T$, which contains additional regularization term(s) for their continual learning approach. In our results section, we demonstrate that even though the inner optimization does not follow the exact optimization of the victim, the constructed noise is very effective against various continual learning approaches.

The bi-level optimization in \eqref{eq:reckless} solely focuses on maximizing forgetting on tasks $1$ through $T-1$, even at the cost of not learning task $T$, i.e., the `reckless threat model'. 
In the `cautious threat model', on the other hand, the attacker must be more strategic, as the victim actively monitors the validation accuracy of the current task. This necessitates a more nuanced approach, where the attacker carefully crafts the training-time noise to maximize forgetting on previous tasks and minimize the error on the clean data from task $T$. 
In other words, this scenario requires balancing the adversarial goal with the need to maintain the model's validation accuracy, making it a significantly more challenging problem. The bi-level optimization for the `cautious' attacker is very similar to that of the `reckless' attacker, with a minor difference in the outer optimization loop:
\begin{align}
\{\delta^i_T\}_{i=1}^{N_T} = \argmax_{\{\delta^i\}_{i=1}^{N_T}} \sum_{t=1}^{T-1}\sum_{j=1}^{M}&\mathcal{L}(\hat{x}^j_t, \hat{y}^j_t, \tilde{\theta}(\delta), \psi^*_t)  -\eta\sum_{i=1}^{N_T} \mathcal{L}(x_T^i,y^i_T,\tilde{\theta}(\delta),\tilde{\psi}_T(\delta))\nonumber\\
\text{s.t. } \quad & \tilde{\theta}(\delta), \tilde{\psi}_T(\delta) = \argmin_{\theta, \psi_T} \sum_{i=1}^{N_T}\mathcal{L}(x^i_T + \delta^i, y^i_T, \theta, \psi_T) \nonumber \\
&~~~~~~~~\ell_{\infty} (\delta^i) < \epsilon,~~\forall i
\label{eq:cautious}
\end{align}
Note that the added loss term with weight $\eta>0$ in the outer-level optimization is designed to ensure that the model trained on the poisoned data performs well on the clean data. Next, we discuss our strategy for solving these bi-level optimizations. 


\subsection{First-Order Approximation}

Solving the bi-level optimization problems in \eqref{eq:reckless} and \eqref{eq:cautious} is intractable, as they require backpropagation through the entire Stochastic Gradient Decent (SGD) training procedure of the inner optimization. Therefore, these bi-level objectives have to be approximated. Following a similar approach to  \cite{huang2020metapoison}, we leverage a first-order method that approximates the bi-level optimization problems using meta-learning \cite{finn2017mamal}.

In short, we simplify the inner objective (i.e., fine-tuning on poisoned data) by limiting the training to only $k$ SGD steps for each evaluation of the outer objective. In other words, the outer backpropagation is only performed through the inner optimization's $k$ unrolled SGD steps. Notably, such $k$-step methods are shown to decrease approximation error exponentially \cite{shaban2019truncated} and have significant generalization benefits \cite{franceschi2018bilevel}. In all our experiments, we set $k=1$. Note that, for each iteration of the inner optimization, the head parameters for task $T$, $\psi_T$, are initialized randomly at each iteration while the backbone parameters, $\theta$, are initialized from the learned backbone at the end of task $T-1$, $\theta^*_{1:T-1}$.

\section{Experiments}

This section provides our experimental results evaluating our proposed attack on various regularization-based multi-head CL algorithms on three benchmark datasets. 

\subsection{Datasets and Model} 
We perform extensive studies on three major continual learning benchmarks: 
\begin{itemize}
    \item \textbf{10-Split CIFAR-100:} CIFAR-100\cite{cifar100} consists of 100 classes of $32 \times 32$ images. We generated 10 ten-way classification tasks by splitting the classes. This dataset serves as our small-scale benchmark.
    \item \textbf{10-Split miniImagenet:} we also evaluated our noise on  miniImageNet\cite{miniImageNet} which is a dataset of 60,000 $84 \times 84$ images, divided in 100 categories. Similarly, we divide the miniImagenet to 10 classification tasks. This dataset serves as our medium-scale benchmark.
    \item \textbf{20-Split tinyImagenet:} As our large-scale benchmark,  we used the 20-split tinyImageNet\cite{tinyIN}, which is a dataset of 200 classes with 100,000 images in total, each with the size of $64 \times 64$. 
\end{itemize}
In all our experiments, we used the ResNet-18 \cite{he2016residual} architecture with Stochastic Gradient Descent (SGD) optimizer with learning rate 1e-2 and batchsize 16. All images were normalized in the range of $[0, 1]$, and the poisoned data was truncated to this range. 

Next, we describe the continual learning algorithms we considered in our experiments.

    

\setlength{\tabcolsep}{3pt}
\begin{table*}[!t]
\centering
\hspace*{-8pt}\begin{tabular}{c?c?c?ccc?ccc}

\Xhline{2\arrayrulewidth}
&&& \multicolumn{3}{c?}{$\bm{\epsilon = 0.1}$} & \multicolumn{3}{c}{$\bm{\epsilon = 0.3}$} \Tstrut\\
&&&&&&&&\\
\textbf{Dataset} & \textbf{Method} & \shortstack{\textbf{Clean} \\ \\ \textsmaller[2]{BWT(Acc)}}& \shortstack{\textbf{Uniform} \\\\ \textsmaller[2]{BWT(Acc)}} & \shortstack{\textbf{Cautious} \\\\ \textsmaller[2]{BWT(Acc)}} & \shortstack{\textbf{Reckless} \\\\ \textsmaller[2]{BWT(Acc)}} &  \shortstack{\textbf{Uniform} \\\\ \textsmaller[2]{BWT(Acc)}} & \shortstack{\textbf{Cautious} \\\\ \textsmaller[2]{BWT(Acc)}} & \shortstack{\textbf{Reckless} \\\\ \textsmaller[2]{BWT(Acc)}} \TBstrut\\
\hline
\hline
\multirow{1.75}{*}{\shortstack{ \\\\ \rotatebox[origin=t]{90}{\textbf{\shortstack{CIFAR-100} }}}}& EWC \cite{kirkpatrick2017overcoming} & -5.2 (68.3)& -5.1 (67.0)& -9.5 (58.8)& -12.6 (51.0)& -12.2 (57.5)& -24.7 (42.2)& -29.1 (25.5)\Tstrut\\
& AFEC \cite{wang2021afec} & -2.9 (65.6)& -3.6 (64.1)& -7.9 (57.2)& -8.8 (55.4)& -14.4 (52.4)& -24.2 (49.0)& -24.6 (36.9)\Tstrut\\
& ANCL \cite{kim2023ancl} & -0.1 (81.5)& -0.2 (80.4)& -0.9 (61.7)& -4.4 (64.9)& -3.6 (68.5)& -6.2 (53.8)& -30.4 (44.3)\Tstrut\\
& MAS \cite{aljundi2017memory} & -1.8 (62.6)& -1.8 (67.7)& -5.8 (61.4)& -6.4 (53.7)& -9.6 (67.5)& -22.8 (50.7)& -17.9 (45.0)\Tstrut\\
& RWALK \cite{chaudhry2018rwalk} & -6.0 (70.1)& -5.1 (68.5)& -16.3 (55.8)& -14.5 (62.8)& -25.5 (48.8)& -32.5 (47.0)& -21.6 (53.5)\Tstrut\\
\hline
\multirow{1.75}{*}{\shortstack{ \\\\\\ \rotatebox[origin=t]{90}{\textbf{ \shortstack{mini \\ ImageNet} }}}}& EWC & -3.9 (56.8)& -1.5 (64.2)& -15.0 (42.5)& -23.1 (28.3)& -14.6 (58.0)& -27.9 (32.2)& -34.4 (22.5)\Tstrut\\
& AFEC & -1.3 (53.3)& -1.4 (52.9)& -14.7 (39.7)& -22.6 (30.9)& -15.1 (37.9)& -27.6 (22.8)& -38.2 (13.2)\Tstrut\\
& ANCL & -1.8 (74.5)& -2.2 (68.5)& -6.7 (34.8)& -5.3 (29.5)& -7.3 (59.0)& -14.6 (38.0)& -14.1 (21.4)\Tstrut\\
& MAS & -6.7 (54.6)& -6.9 (57.2)& -25.7 (40.3)& -30.3 (23.6)& -18.8 (48.8)& -39.8 (22.6)& -38.4 (16.4)\Tstrut\\
& RWALK & -5.6 (66.3)& -8.4 (53.6)& -13.5 (45.9)& -17.9 (38.0)& -22.6 (38.0)& -21.4 (37.4)& -27.4 (22.4)\Tstrut\\
\hline
\multirow{1.75}{*}{\shortstack{ \\\\\\ \rotatebox[origin=t]{90}{\textbf{ \shortstack{ tiny\\ImageNet} }}}}& EWC & -0.4 (53.0)& 0.7 (52.6)& -5.8 (39.2)& -7.1 (35.6)& -7.3 (44.6)& -28.4 (11.6)& -25.9 (16.2)\Tstrut\\
& AFEC & -1.3 (51.6)& -2.8 (51.0)& -10.3 (34.2)& -15.4 (30.4)& -13.5 (34.0)& -26.2 (15.0)& -27.8 (14.8)\Tstrut\\
& ANCL & -1.7 (73.4)& -1.5 (72.2)& -4.3 (46.0)& -6.7 (32.6)& -3.4 (51.8)& -10.7 (37.2)& -16.0 (22.8)\Tstrut\\
& MAS & -1.1 (59.4)& -1.8 (60.0)& -5.0 (54.2)& -12.9 (31.2)& -8.0 (46.8)& -25.5 (37.8)& -28.0 (22.6)\Tstrut\\
& RWALK & -14.1 (40.6)& -14.6 (39.6)& -25.9 (38.8)& -25.7 (44.2)& -29.8 (21.2)& -33.5 (26.6)& -33.3 (25.8)\Tstrut\\
\Xhline{2\arrayrulewidth}
\end{tabular}
\caption{Results of our attack on different datasets and different methods. We reported the results using two different $\ell_\infty$ for various noises. The clean column corresponds to regular learning of the last task without adding the BrainWash. Uniform stands for uniform noise in the range of $[-\epsilon, +\epsilon]$.}
\label{tab:main}
\end{table*}

\subsection{CL Methods}
In our experiments, we consider 5 renowned regularization-based methods starting from the classic Elastic Weight Consolidation (EWC) \cite{kirkpatrick2017overcoming}, Memory Aware Synapsis (MAS) \cite{aljundi2017memory}, and Riemannian Walk (RWALK), to more recent methods like Active Forgetting of Negative Transfer (AFEC) \cite{wang2021afec} and Auxiliary Networks in
Continual Learning (ANCL) \cite{kim2023ancl}. Generally, the regularization-based CL methods assign importance values to network parameters and penalize the training for drastic changes in the important parameters. At a high level, this can be formulated as:
\begin{align}
    \theta^*_{1:T} = \argmin_\theta \sum_{i=1}^{N_T} \mathcal{L}(x_T^i,y_T^i,\theta)+\lambda \mathcal{R}_{\text{CL}}(\theta,\theta_{1:T-1}^*),   
\end{align}
where $\mathcal{R}_{\text{CL}}$ is a CL method-dependent regularizer that enforces stability of the continual learner, $\lambda$ is the regularization coefficient that balances the stability vs. plasticity (or forgetting vs. intransigence) trade-off, and $\theta$ is initialized at $\theta^*_{1:T-1}$.
It is widely accepted that the performance of regularization-based methods highly depends on the choice of $\lambda$. 

\subsection{Evaluation Metrics}

For our evaluation metric, we use the Backward Transfer (BWT) \cite{lopez2017gradient} and the poisoned model's accuracy on the last task. For the sake of completion, let $A_{t,i}$ denote the performance of the CL model on task $i$ after learning task $t>i$. Then, BWT is defined as: 
\begin{align}
    \text{BWT} = \frac{1}{T-1}\sum_{i=1}^{T-1} A_{T,i}-A_{i,i}.
\end{align}
Importantly, maximizing forgetting on previous tasks coincides with minimizing BWT. Hence, our poisoning attack is expected to decrease BWT.

\subsection{Experiment Setup}\label{sec:exp_setup}


We explore regularization-based CL models that have been trained on \(T-1\) tasks using the ideal regularization coefficient (\(\lambda\)) for their respective CL methods, aiming for a balance between plasticity and stability. For each victim model, we introduce poison to task \(T\) under two different \(\ell_\infty\) norm bounds: \(\epsilon=0.1\) and \(\epsilon=0.3\). These bounds are applied in both 'reckless' and 'cautious' attacker scenarios, as described by Equations \eqref{eq:reckless} and \eqref{eq:cautious}, leading to four distinct experimental setups. The victim models then learn the poisoned task \(T\) using their CL methods. Post learning, we evaluate the Backward Transfer (BWT) and the accuracy on the clean data of the last task for these victim models. For comparative analysis, we also include the BWT and accuracy of the victim models trained on the unpoisoned version of task \(T\) and on task \(T\) with added uniform noise. The outcomes of all these experimental configurations are detailed in Table \ref{tab:main}. Our results indicate a significant BWT decrease when models are trained on BrainWash data. Additionally, it is observed that the `cautious' attacker often achieves higher accuracy on task \(T\) compared to the `reckless' attacker, albeit with a trade-off of a less potent attack. As anticipated, the poisoning effect with \(\epsilon=0.3\) is markedly more substantial than that with \(\epsilon=0.1\).



\begin{figure}
    \centering
    \includegraphics[width=0.5\columnwidth]{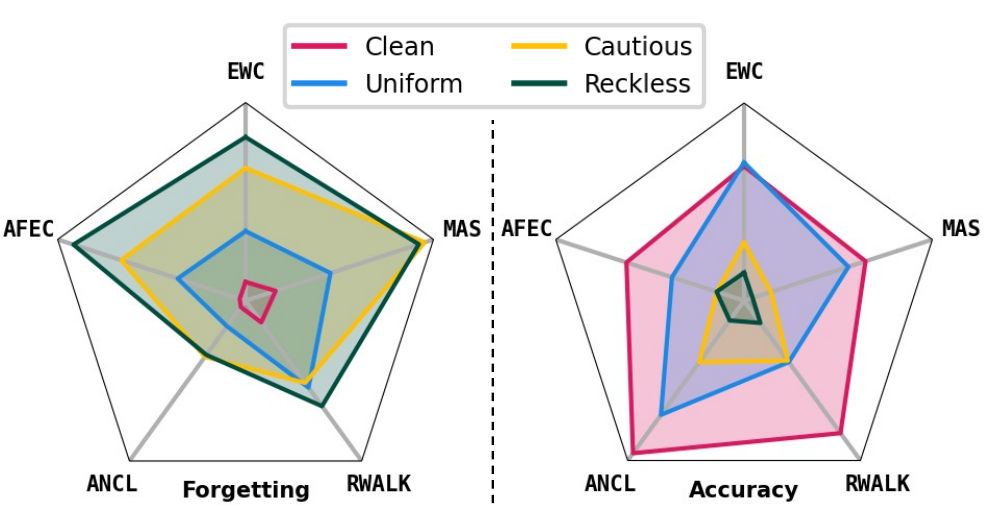}
    \caption{Forgetting (i.e., negative backward transfer) and accuracy of task $T$ for different attacking strategies and on different CL approaches trained on miniImageNet with $\epsilon=0.3$. As can be seen, forgetting is minimal when the continual learner is trained on the clean data. Adding uniform noise increases forgetting, while `cautious' and `reckless' attackers increase forgetting by a large margin. Also, the trade-off between the attack's success and the accuracy of the last task is apparent.}
    \label{fig:spider}
\end{figure}

To aid in comprehending the results presented in Table \ref{tab:main}, we have depicted the miniImageNet results as a spider chart in Figure \ref{fig:spider}. Key observations from this visualization include: 1) a discernible trade-off between enhanced forgetting (i.e., decrease in BWT) and improved accuracy on the last task, 2) a consistent increase in last task accuracy for the `cautious' attacker, though this comes with a reduction in forgetting efficiency, and 3) the notable superiority of ANCL and RWALK in withstanding our poisoning attack compared to other evaluated methods.

Next, we conduct various ablation studies to gain deeper insights into BrainWash.

\begin{figure}[b]
    \begin{minipage}[t]{.45\textwidth}
    \centering
    
    \includegraphics[width=1\textwidth]{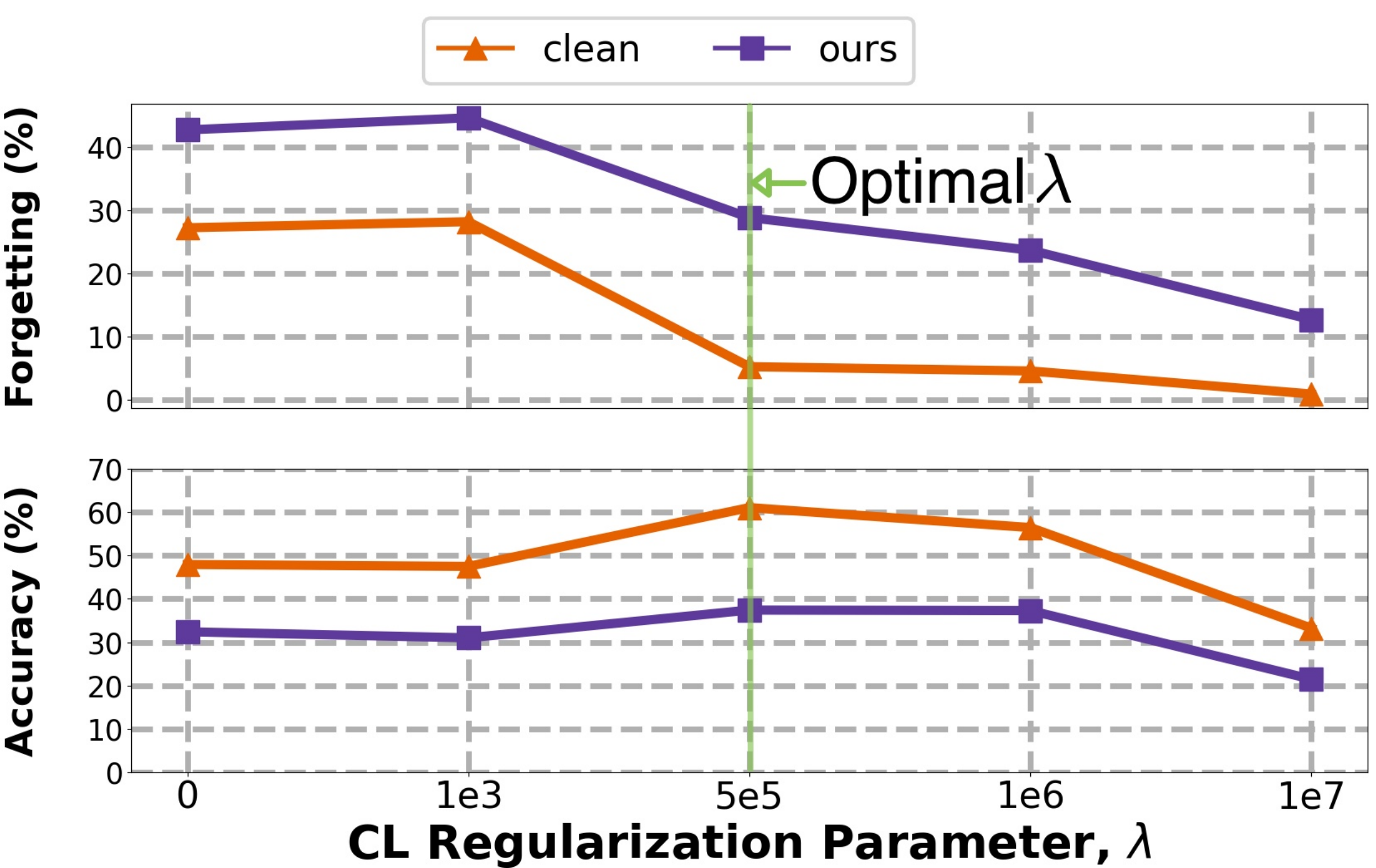}
    \vspace*{1mm}
    \caption{Performance of BrainWash against victims using different regularization coefficients.}
    \label{fig:ablation_lamb_mix}
    \end{minipage}\hfill
    \begin{minipage}[t]{.45\textwidth}
    \centering
    \includegraphics[width=1\textwidth]{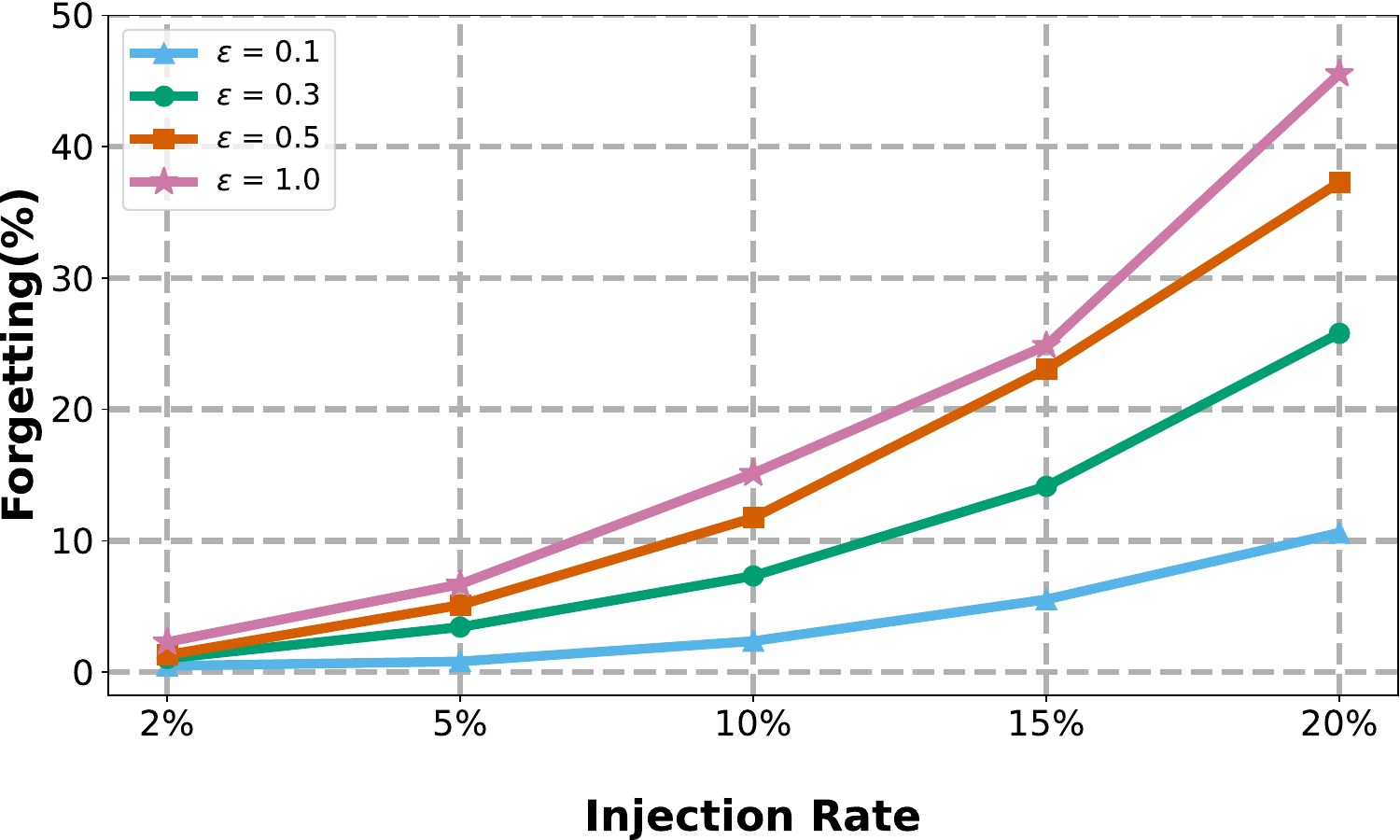}
    \caption{Effect of different injection ratio on the effectiveness of the attack. Forgetting is defined as negative value of backward transfer. 
    Note that the x-axis reports the injection rate for the whole dataset, so $20\%$ means $100\%$ of the final task is poisoned.}
    \label{fig:inj_abl}
    \end{minipage}
\end{figure}



\section{Abaltion Studies}
We performed various ablation studies to evaluate the sensitivity of BrainWash to different design choices. Please note that throughout the ablation experiments, we use the term \textbf{``forgetting"}, corresponding to the negative of BWT. Moreover, all our ablation studies were performed on EWC. 

\subsection{Sensitivity to \large$\lambda$} \label{sec:ablation_lamb}

As previously mentioned in Section \ref{sec:exp_setup}, our results reflect the victim model's natural behavior, particularly in choosing the optimal \(\lambda\) for their CL algorithm. It's important to note that the degree of induced forgetting and the overall performance of the victim is significantly influenced by the value of \(\lambda\). Also, there is a notion that increasing the network's stability (i.e., opting for higher values of \(\lambda\)) might act as an effective defense against BrainWash, under the assumption that the lesser the amount of "bad data" learned by the victim, the lower the level of forgetting. This section delves into the relationship between BrainWash's effectiveness and the choice of \(\lambda\).

To explore this, we conducted an experiment on 10-split CIFAR-100. Here, the assumption is that the victim model has been trained on the first nine tasks using various fixed \(\lambda\) values, which remain constant throughout the CL process. The BrainWash is then applied to the data of the \(10^{th}\) task. It is crucial to point out that BrainWash remains oblivious to the \(\lambda\) value used by the victim. Fig. \ref{fig:ablation_lamb_mix} shows the sensitivity of BrainWash to the victim's choice of $\lambda$, which identifies different degrees of network plasticity. The top plot shows the amount of forgetting while the bottom plot demonstrates the average accuracy of the victim on the past $9$ tasks as a function of $\lambda$. Figure \ref{fig:ablation_lamb_mix} indicates that BrainWash consistently leads to increased forgetting and reduced average performance across different \(\lambda\) values, even when the victim opts for the optimal \(\lambda\) for their context, such as \(\lambda=5e5\) in this experiment. Although increasing the network's intransigence (using higher \(\lambda\) values) marginally diminishes the attack's effectiveness, this approach also significantly compromises the victim's overall performance. Therefore, BrainWash proves to be resilient to varying \(\lambda\) choices, and utilizing high \(\lambda\) values does not constitute an efficient defense strategy.

\subsection{Dependency on Injection Rate and Noise Norm}

To delve deeper into the dynamics of BrainWash, we assessed its impact by varying the noise injection rate (the percentage of data that is poisoned) and the noise magnitude. In an experiment using CIFAR-100 divided into five tasks, we poisoned the last task with different injection rates and noise magnitudes. It's important to note that poisoning the entire task equates to a $20\%$ injection rate (since one out of the five tasks is poisoned), and poisoning $10\%$ of the last task corresponds to a $2\%$ injection ratio. 

Figure \ref{fig:inj_abl} demonstrates how both the injection rate and noise amplitude influence the extent of for getting. Our observations indicate a direct correlation between forgetting, noise norm, and injection rate. This figure also reveals a constant trade-off between the subtlety of the noise and the amount of forgetting induced: increasing either the norm or the injection rate leads to more pronounced forgetting. However, in scenarios where stealthiness is crucial, such increases in noise or rate can potentially expose BrainWash. Despite this, the results show that effective forgetting can still be achieved even with a minimal injection rate.

\subsection{The Effect of Model Inversion}

As previously mentioned, we considered a scenario where the attacker might not have access to data from previous tasks. To address this, we suggested using model inversion, employing inverted samples as proxies. In this context, we examined the significance of model inversion for the effectiveness of BrainWash. Our study on CIFAR100, segmented into 10 tasks, evaluated the impact of BrainWash under three distinct conditions: 1) access to actual data from preceding tasks, 2) application of a basic model inversion technique without any regularization, and 3) utilization of regularized model inversion, as detailed in Section \ref{sec:inversion}.

The findings are presented in Table \ref{tab:data_proxy}, which illustrates the percentage of forgetting associated with each of these strategies. The term `Clean' refers to the inherent forgetting experienced by the continual learner trained on the clean final task. Notably, BrainWash, when implemented with regularized model inversion, achieves results comparable to the scenario with direct access to real data. Furthermore, the robustness of BrainWash to the choice of model inversion method is evident, as the basic, non-regularized inversion demonstrates only marginal underperformance compared to its more advanced counterpart. 

\setlength{\tabcolsep}{3pt}
\begin{table}[t]
    \centering
    
    \begin{tabular}{c|cccc}
    \Xhline{2\arrayrulewidth}
        
         &  \small \textbf{Clean}  & \smaller\shortstack{\\ \textbf{Inv Data} \\ \textbf{No Reg}} & \smaller\shortstack{\\ \textbf{Inv Data} \\ \textbf{with Reg}}& \smaller\shortstack{\\ \textbf{Real} \\ \textbf{Data}}\\
         \Xhline{2\arrayrulewidth}
         \textbf{Forgetting} & 5.2 & 23.6 & 28.8 & 28.16 \\
    
         \Xhline{2\arrayrulewidth}
    \end{tabular}
    \vspace{4mm}
    \caption{Difference between the induced forgetting while using different alternatives for the past data}
    \label{tab:data_proxy}
\end{table}

\setlength{\tabcolsep}{5pt}
\begin{table}[b]
    \centering
    
    \begin{tabular}{c|ccc}
    \Xhline{2\arrayrulewidth}
        
         \textbf{Method}&  \textbf{10 \small Tasks}  & \textbf{15 \small Tasks}& \textbf{20 \small Tasks}\\
         \Xhline{2\arrayrulewidth}
         Clean & 1.4&2.17&2.16\\
         Ours & 17.47&14.61&14.04 \\
         \Xhline{2\arrayrulewidth}
    \end{tabular}
    \vspace{4mm}
    \caption{Effect of different number of tasks on forgetting}
    \vspace{-.1in}
    \label{tab:num_tasks}
\end{table}

\subsection{Different Task Lengths}

In our final ablation study, we focused on assessing the efficacy of BrainWash at various stages of a continual learner's training. We divided CIFAR-100 into 20 five-way classification tasks and measured the extent of forgetting immediately after introducing noise in the \(10^{th}\), \(15^{th}\), and \(20^{th}\) tasks. The results, as depicted in Table \ref{tab:num_tasks}, confirm that BrainWash effectively induces forgetting at different training stages of the continual learner. 

An intriguing finding is the variation in the injection rate of BrainWash across these stages. For instance, at task 10, BrainWash has an injection rate of $10\%$ (being applied to one out of ten tasks). However, this rate decreases when poisoning is applied to 15 or 20 tasks, leading to a corresponding reduction in the attack's strength. This observed decline in potency illustrates how the impact of BrainWash is influenced by its relative scale in the context of the overall training process.


\section{Conclusion}

This study presents BrainWash, an innovative poisoning attack for regularization-based continual learning (CL) models. Its primary objective is to maximize the forgetting of a continual learner on previously learned tasks. We introduced two threat models: the `reckless' and the `cautious' attacker. Both threat models assume that the attacker can only access the trained model and the clean data from the last task. Critically, the attacker is unaware of the specific CL method employed by the victim or any related hyperparameters. Our core strategy involves using model inversion to approximate data from earlier tasks. The attacker then employs a bilevel optimization problem to poison the current task's data. When the victim trains their model on this manipulated data, their performance on prior tasks is adversely affected. The `reckless' attacker disregards the victim's performance on the last task's clean data, while the `cautious' attacker seeks to preserve high accuracy on it.

In our extensive experiments, we employed five well-known continual learning methods and three benchmark datasets to demonstrate the efficacy of BrainWash as a potent poisoning attack. Moreover, we provided a series of detailed ablation studies to offer a thorough understanding of BrainWash's mechanics and impacts.

\bibliographystyle{unsrtnat}
\bibliography{main_arxiv}  






\end{document}